%% file: main.tex
\documentclass[10pt,twocolumn,letterpaper]{article}

\usepackage{wacv}
\makeatletter
\@namedef
{ver@everyshi.sty}{}
\makeatother

\usepackage{times}
\usepackage{epsfig}

\usepackage{graphicx}
\usepackage{amsmath,bm}
\usepackage{amssymb}
\usepackage{booktabs}
\usepackage{color, colortbl}
\usepackage{booktabs}
\usepackage[percent]{overpic}
\usepackage{xcolor}
\usepackage{verbatim}
\usepackage{multirow}
\usepackage{highlight}

\usepackage{subcaption}
\usepackage{balance}

\newcolumntype{;}{!{\vrule width 2pt}}

\def\wacvPaperID{1524} 

\wacvfinalcopy 

\ifwacvfinal
\def\assignedStartPage{9876} 
\fi

\wacvapplicationstrack

\ifwacvfinal
\usepackage[colorlinks,breaklinks=true,bookmarks=false]{hyperref}
\else
\usepackage[pagebackref=true,breaklinks=true,colorlinks,bookmarks=false]{hyperref}
\fi

\usepackage[capitalize]{cleveref}
\crefname{section}{Sec.}{Secs.}
\Crefname{section}{Section}{Sections}
\Crefname{table}{Table}{Tables}
\crefname{table}{Tab.}{Tabs.}

\newcommand\blfootnote[1]{%
  \begingroup
  \renewcommand\thefootnote{}\footnote{#1}%
  \addtocounter{footnote}{-1}%
  \endgroup
}

\ifwacvfinal
\else
\fi

\definecolor{somegray}{rgb}{0.5, 0.5, 0.5}
\newcommand{\darkgrayed}[1]{\textcolor{somegray}{#1}}
\makeatletter
\newcommand*\titleheader[1]{\gdef\@titleheader{#1}}
\AtBeginDocument{%
  \let\st@red@title\@title
  \def\@title{%
    \vskip-3em
    \bgroup\normalfont\large\centering\@titleheader\par\egroup
    \vskip1.5em\st@red@title}
}
\makeatother

\titleheader{\darkgrayed{This paper has been accepted for publication at the \\
IEEE/CVF Winter Conference on Applications of Computer Vision (WACV), Waikoloa, 2023.
\copyright IEEE}}

\title{ScanNeRF: a Scalable Benchmark for Neural Radiance Fields}

\author{Luca De Luigi$^{2 *}$ \quad Damiano Bolognini$^{1 *}$ \quad Federico Domeniconi$^{1 *}$ \\%
Daniele De Gregorio$^1$ \quad Matteo Poggi$^2$ \quad Luigi Di Stefano$^{1,2}$ %
\\
{\small $^1$Eyecan.ai \quad $^2$University of Bologna} 
\\
{\small \tt {Project page: \url{https://eyecan-ai.github.io/scannerf/}}}\\
}

\begin{document}

\twocolumn[{
\renewcommand\twocolumn[1][]{#1}
\maketitle
\begin{center}
    \vspace{-0.5cm}
    \begin{tabular}{c}
        \begin{overpic}[height=0.28\textwidth]{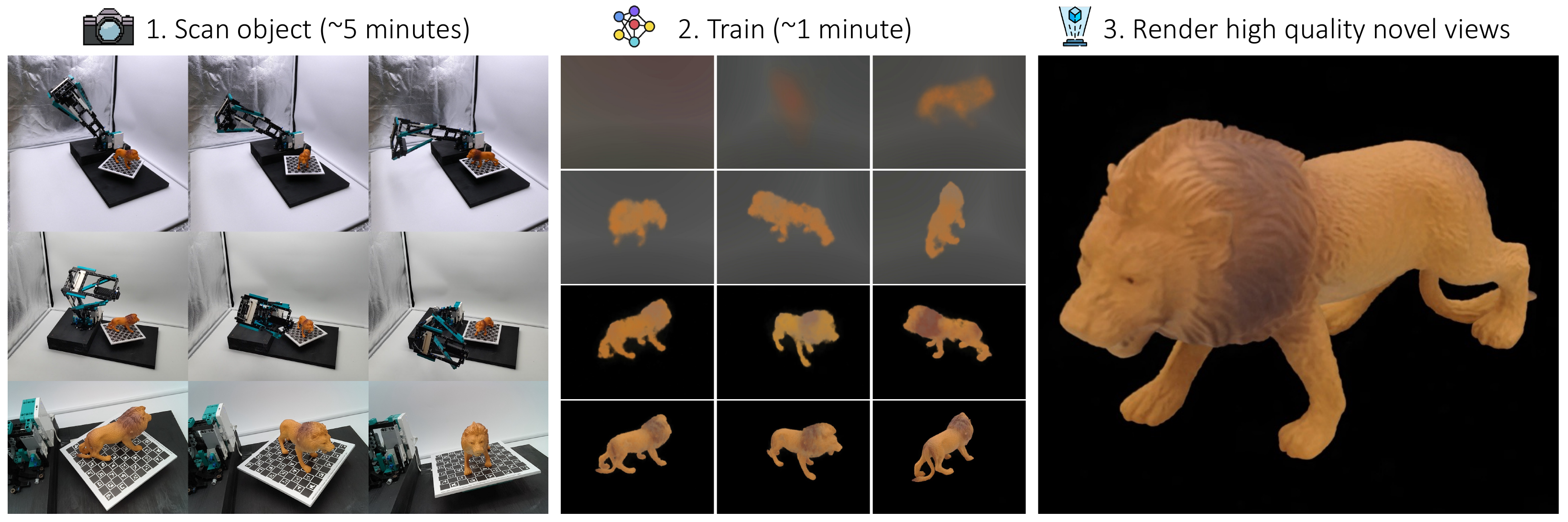}
        \end{overpic} \\
    \end{tabular}
\end{center}
\vspace{-0.25cm}
\small \hypertarget{fig:teaser}{Figure 1.} \textbf{Overview of the ScanNeRF framework.} Our scan station (left) allows for collecting thousands of images of an object in a few minutes. Then, modern NeRF variants \cite{sun2021direct,yu2021plenoxels,mueller2022instant} can be trained on them in few minutes (center), producing a digital twin of the object itself and allowing for high-quality, novel view synthesis of it (right).
\vspace{0.2cm}
}]
\thispagestyle{empty}

\begin{abstract}

In this paper, we propose the first-ever real benchmark thought for evaluating Neural Radiance Fields (NeRFs) and, in general, Neural Rendering (NR) frameworks.
We design and implement an effective pipeline for scanning real objects in quantity and effortlessly. Our \textit{scan station} is built with less than 500\$ hardware budget and can collect roughly 4000 images of a  scanned object in just 5 minutes.
Such a platform is used to build \textit{ScanNeRF}, a dataset characterized by several train/val/test splits aimed at benchmarking the performance of modern NeRF methods under different conditions.
Accordingly, we evaluate three cutting-edge NeRF variants on it to highlight their strengths and weaknesses.
The dataset is available on our project page, together with an online benchmark to foster the development of better and better NeRFs.

\end{abstract}

\section{Introduction}
\blfootnote{{$^*$joint first authorship}}What is the Metaverse? \emph{Stephenson} coined this portmanteau in his novel \emph{Snow Crash}, hypothesizing that in the 21st century humans, thanks to goggles, would be immersed in virtual worlds mixed with real ones. And here we are!
At the time, however, the technology to realize the Metaverse was still hypothetical, but today Cross Reality (XR or Extended Reality) is a fact. XR is made up of a multitude of technologies and variants, such as Virtual Reality and Augmented Reality, but they all share a single paradigm: seamless interaction between virtual environments, digital objects and people. That is the Metaverse! But it does not exist yet, and all that is Digital is often only a virtual representation of the real world. How much will it cost us, then, to transport all our real world into the virtual one?

For Computer Vision and Computer Graphics experts, it is clear what it means to transport an object from the real world to the virtual world: a 3D reconstruction! But 3D reconstructions are expensive, slow, and not all types of objects can be digitized. Yet today, thanks to Deep Learning, we have another way to teleport objects into the Metaverse: Neural Rendering \cite{tewari2021advances}. The basic idea is simple: why reconstructing an object in 3D if we have to render it back in 2D to visualize it by a VR / AR viewer? Neural Rendering (NR) allows us to ask a neural network ``render this object from this point of view'', et voilà! Moreover, some of the state-of-the-art NR approaches -- e.g., Neural Radiance Fields (NeRFs) \cite{MildenhallSTBRN20} -- allow us to deploy  a simple MLP to represent an entire scene (or object), reducing the spatial cost of a digital object from Gigabytes to a few Kilobytes.

In this paper, we will focus on one key aspect: the gate to the Metaverse. We have built an effective object scanning station, dubbed \emph{ScanNeRF}, which allows for  generating ready-to-use data to train and evaluate state-of-the-art Neural Radiance Fields techniques. Using this efficient and simple scanning system, we generated the first real dataset with high quality images, pixel masked objects, controlled and repeatable camera poses,  specifically designed to evaluate NeRFs. Firstly, this  allows us to realize a benchmark for research in the area of Neural Rendering. Secondly, it enables to formally describe which and how many views are best for generating a virtual representation of an object, as well as to unveil some intriguing challenges for the future -- e.g., how to fully render an object from any viewpoint, given images mostly collected from a single side of it.

To the best of our knowledge, our work is the first to show that with a simple hardware, made of LEGO, and a low budget -- less than 500\$ -- it is possible to build Digital Twin from \textbf{real} objects, rather than focusing on synthetic ones   as in most of NeRF papers \cite{MildenhallSTBRN20}.

Our contributions are as follows:

\begin{itemize}
    \item We present a simple, yet effective platform for collecting thousands of images to train NeRFs, or in general, NR frameworks.
    
    \item We release a novel benchmark, ScanNeRF, featuring thousands of images depicting real objects collected, in inward-facing setting. 
    
    \item For each object in the benchmark, we define a multitude of training/validation/testing splits in order to study different properties and stress the performance of NeRF variants. Moreover, we evaluate the performance of three modern NeRFs on these splits, to highlight their strengths and weaknesses under different experimental settings. 
\end{itemize}

Fig. \hyperlink{fig:teaser}{1} presents an overview of our ScanNeRF framework. Fitting a NeRF on the scanned object produces a \textit{digest} of it, ready to be transported into the Metaverse. {Actually, this representation is very different from that of a classical Digital Twin, this is indeed a Neural Twin®.}

\section{Related work}

We review the literature on Neural Radiance Fields\footnote{A curated list of NeRF papers is constantly updated at \url{https://github.com/yenchenlin/awesome-NeRF}}, representing the most relevant topic for our work.

\textbf{Neural radiance fields.} NeRF~\cite{MildenhallSTBRN20} represents nowadays the most popular paradigm for novel view synthesis, rapidly conquering the main stage over explicit approaches exploiting CNNs \cite{ZhouTFFS18,FlynnBDDFOST19,MildenhallSCKRN19,SrinivasanTBRNS19,LiXDS20,TuckerS20,LombardiSSSLS19,SitzmannTHNWZ19,HeCJS20}.
Peculiar to NeRF is an implicit, continuous volumetric representation encoded by a multilayer perceptron (MLP) -- opposed to discrete representations such as voxel grids or multi-plane images -- which enables  to retrieve color and density of queried 3D points and to render images through differentiable ray casting.
Vanilla NeRF has been rapidly extended to deal with different setups, e.g.,  relighting~\cite{SrinivasanDZTMB21,ZhangSDDFB21,BossBJBLL21}, deformable agents~\cite{ParkSBBGSM21,TretschkTGZLT21,GafniTZN21,NoguchiSLH21,ParkSHBBGMS21}, dynamic scenes~\cite{Martin-BruallaR21,PumarolaCPM21,LiNSW21,XianHK021,GaoSKH21}, multi-resolution images \cite{BarronMTHMS21} or to implement generative models~\cite{SchwarzLN020,ChanMK0W21,KosiorekSZMSMR21}.

Despite the elegant formulation and impressive quality of the synthesised views, the original NeRF suffers of some notable limitations, such as, in particular, the long training process -- a few days in its very first implementation \cite{MildenhallSTBRN20} -- together with the requirement to perform a standalone training from scratch for any new scene and the slow rendering speed -- definitely far from real-time.

\textbf{Faster training.} Speeding-up the training procedure represents the main barrier to break in order to deploy NeRF in real applications, as it would soften the limitation of requiring a scene-specific training.
The main approaches proposed in literature rely on a pre-training phase~\cite{YuYTK21,SRF,ChenXZZXYS21,WangWGSZBMSF21}, deploy additional depth information estimated by means of Multi-View Stereo (MVS) methods~\cite{LiuPLWWTZW21,DengLZR21}, use neural rays~\cite{LiuPLWWTZW21},  exploit explicit representations~\cite{yu2021plenoxels} or combine them with implicit ones ~\cite{sun2021direct,mueller2022instant}.

\textbf{Faster rendering.} Achieving real-time rendering is highly desirable to improve end-user experience, possibly allowing for interactive visualization of novel viewpoints of a given object.
Recent works exploit octree structures~\cite{LiuGLCT20} to avoid redundant MLP queries in empty space, split a single MLP in thousands of tiny ones~\cite{ReiserPLG2021} or leverage explicit volumetric representations~\cite{WizadwongsaPYS21,YuLTLNK2021,GarbinKJSV2021,HedmanSMBD2021}.

\textbf{Next-generation NeRFs.} At the time of writing, a few very recent works  stand out in terms of both  training and inference speed. DirectVoxGo (DVGO)~\cite{sun2021direct} combines implicit and explicit representations, using voxel grids together with a light MLP. Plenoxels~\cite{yu2021plenoxels} gets rid of the MLP and directly optimizes colors over a voxel grid. Instant Neural Graphic Primitives (Instant-NGP)~\cite{mueller2022instant} makes use of hash tables and optimized MLP implementations. Any of these frameworks can be easily trained in less than 10 minutes and can achieve good rendering speed without  noticeable deterioration of rendering quality. For these reasons, we will train DirectVoxGo, Plenoxels and Instant-NGP as baselines for evaluation within our new ScanNeRF benchmark, since we argue future developments in this field will follow this direction.

\begin{table}[]
    \centering
    \scalebox{0.65}{
    \begin{tabular}{l|l|lllll}
        & & \# Total & \# Images & Train & Test & Withheld\\
        Dataset & Type & scenes & per scene & splits & splits & images\\
        \hline
        NeRF Blended \cite{MildenhallSTBRN20} & Synth. & 8 & 300 & 1 & 1 & No\\
        BlendedMVG \cite{YaoLLZRZFQ20} & Synth. & 508 & 200-4\,000 & NA & NA & No\\
        \hline
        LLFF \cite{MildenhallSCKRN19} & Real & 8 & 30 & 1 & 1 & No\\
        DTU \cite{aanaes2016large} & Real & 124 & 49-64 & NA & NA & No\\
        CO3D \cite{reizenstein2021common}& Real & 18\,619 & 100 & 2 & 2 & Yes\footnotemark\\
        ScanNet \cite{dai2017scannet} & Real & 1613 & 500-5\,000 & NA & NA & No\\
        Tanks \& Temples \cite{KnapitschPZK17} & Real & 14 & 4\,000-20\,000 & NA & NA & No \\
        \hline
        \textbf{ScanNeRF (ours)} & \textbf{Real} & \textbf{35} & \textbf{4\,000} & \textbf{12} & \textbf{9} & \textbf{Yes} \\
        \hline
    \end{tabular}}
    \caption{\textbf{Comparison between datasets.} We report properties of existing datasets and our ScanNeRF benchmark.} 
    \label{tab:datasets}
\end{table}

\textbf{Datasets.} NeRF and follow-up implementations are usually evaluated on a few, established benchmarks belonging to two acquisition settings, namely  forward-facing and inward-facing, 
The most popular benchmarks are  NeRF blender \cite{MildenhallSTBRN20}, made of 8 synthetic inward-facing scenes with 100 training images and 200 testing images, and LLFF \cite{MildenhallSCKRN19}, consisting of  8 forward-facing scenes counting about 30 images each.
{More recently, MVS datasets such as DTU \cite{aanaes2016large}, Tanks \& Temples \cite{KnapitschPZK17} and BlendedMVG \cite{YaoLLZRZFQ20} have been used for this purpose, together with a few more such as CO3D \cite{reizenstein2021common} and ScanNet \cite{dai2017scannet} collected through extremely time-consuming practises.}

We argue that the aforementioned benchmarks limit the evaluation of NeRF variants under different aspects, since i) some of them \cite{MildenhallSTBRN20,MildenhallSCKRN19,aanaes2016large} provide a few hundred images only, ii) none of them allows for seamlessly scaling the amount of training images or their distribution across the scene {and iii) none explicitly defines a testing set -- i.e., the evaluation is carried out on images available to the researchers, possibly leading to biased results}. In this paper, instead, we implement a framework allowing for scalable data collection of a multitude of scenes. {For each of them, we explicitly define a testing set, made of frames for which only camera poses are made publicly available, while images are withheld to avoid unfair evaluation.} This paves the way towards establishing a next-generation benchmark for research in Neural Radiance Fields and related  techniques. 
\cref{tab:datasets} shows a comparison between the existing datasets introduced before and the proposed ScanNeRF bechmark.

\footnotetext{The possibility to evaluate on withheld images has been added in a second version of the dataset, released concurrently with our work.}

\section{Background on Neural Radiance Fields}

Neural Radiance Fields (NeRF)~\cite{MildenhallSTBRN20} encode a 3D scene into an implicit representation, i.e.,  a function $\bm{F}_0$ mapping any space position $\bm{x}$ and viewing direction $\bm{d}$ pair into density $\sigma$ and view-dependent color emission $\bm{c}$ :
\begin{equation}
    \bm{F}_{0} : (\bm{x}, \bm{d}) \rightarrow (\bm{c}, \sigma).
\end{equation}
Such implicit mapping is learned through a multilayer perceptron (MLP). Specifically, an intermediate MLP$^{\mathrm{(pos)}}$ infers density $\sigma$ alongside an intermediate embedding, $\bm{e}$, used by a shallower MLP$^{\mathrm{(rgb)}}$ together with viewing direction $\bm{d}$ to predict color:

\begin{equation}
\label{eq:nerf_mlp}
\begin{split}
    (\sigma, \bm{e}) &= \operatorname{MLP}^{\mathrm{(pos)}}(\bm{x})~,\\
    \bm{c} &= \operatorname{MLP}^{\mathrm{(rgb)}}(\bm{e}, \bm{d}).
\end{split}
\end{equation}
Before feeding $\bm{x}$ to MLP$^{\mathrm{(pos)}}$, the 3D coordinates are projected into a higher-dimensional space through a positional encoding based on Fourier features \cite{TancikSMFRSRBN20}  which enables to learn to represent more accurately  the high-frequencies of the underlying function. 

\begin{equation}
\label{eq:encoding}
\begin{split}
    \gamma(x) = ( &\sin{(2^0\pi x)}, \cos{(2^0\pi x)}, ... , \\ &\sin{(2^{L-1}\pi x)}, \cos{(2^{L-1}\pi x)}).
\end{split}
\end{equation}

To render an image, i.e., to get color $\hat{C}(\bm{p})$ of any pixel $\bm{p}$, a ray $\bm{r}$ from the camera center through the pixel $\bm{p}$ is cast through the 3D space. 
Then, pixel color $\hat{C}(\bm{p})$ is obtained through volumetric rendering according to the optical model by Max~\cite{Max95a}:
\begin{equation}
\begin{split}
    \hat{C}(\bm{p}) &= \int_{t_n}^{t_f} T(\bm{t}) \sigma(r(\bm{t})) c(r(\bm{t}),\bm{t}) \textit{dt} \\ 
    \quad\quad T(\bm{t}) &= \text{exp}\Big(-{\int_{t_n}^{t}\sigma(r(\bm{s}))\textit{ds}}\Big)
\end{split}
\end{equation}
with $T(\bm{t})$ being the accumulated transmittance along the ray $\bm{t}$ from near plane $t_n$ to any specific point $\bm{t}$. The value of such integral is estimated through quadrature, by sampling $N$ evenly distant 3D points along the ray: 
\begin{equation} \label{eq:volume_rendering}
\begin{split}
    \hat{C}(\bm{r}) &= \left( \sum_{i=1}^{K} T_i \alpha_i \bm{c}_i \right) + T_{\scriptscriptstyle K+1} \bm{c}_{\mathrm{bg}}  ~, \\
    \alpha_i &= \operatorname{alpha}(\sigma_i, \delta_i) = 1 - \exp(-\sigma_i \delta_i) ~, \\ 
    T_i &= \prod_{j=1}^{i-1} (1 - \alpha_j) ~,
\end{split}
\end{equation}
with $\alpha_i$ being the probability of termination at the point $i$, $\delta_i$ the distance to the adjacent sampled point, and $\bm{c}_{bg}$ a pre-defined background color.

Given a set of training images with known camera poses, a NeRF model is trained by minimizing the photometric MSE between the pixel color $C(\bm{r})$ in the training image and the rendered color $\hat{C}(\bm{r})$:
\begin{equation} \label{eq:photo_loss}
    \mathcal{L}_{\mathrm{photo}} = \frac{1}{|\mathcal{R}|} \sum_{r\in\mathcal{R}} \left\|\hat{C}(\bm{r}) - C(\bm{r})\right\|_2^2 ~ ,
\end{equation}
with $\mathcal{R}$ with the set of rays in a single batch.

\section{The ScanNeRF benchmark}

In this section, we describe both hardware and software components of our ScanNeRF framework. We start by introducing our acquisition platform, then we describe the post-processing steps implemented to select the final images and the masking strategy used to extract objects. To conclude, we highlight the overall organization of the produced dataset.

\subsection{Scan station setup}

\begin{figure}
    \centering
    \includegraphics[width=0.45\textwidth]{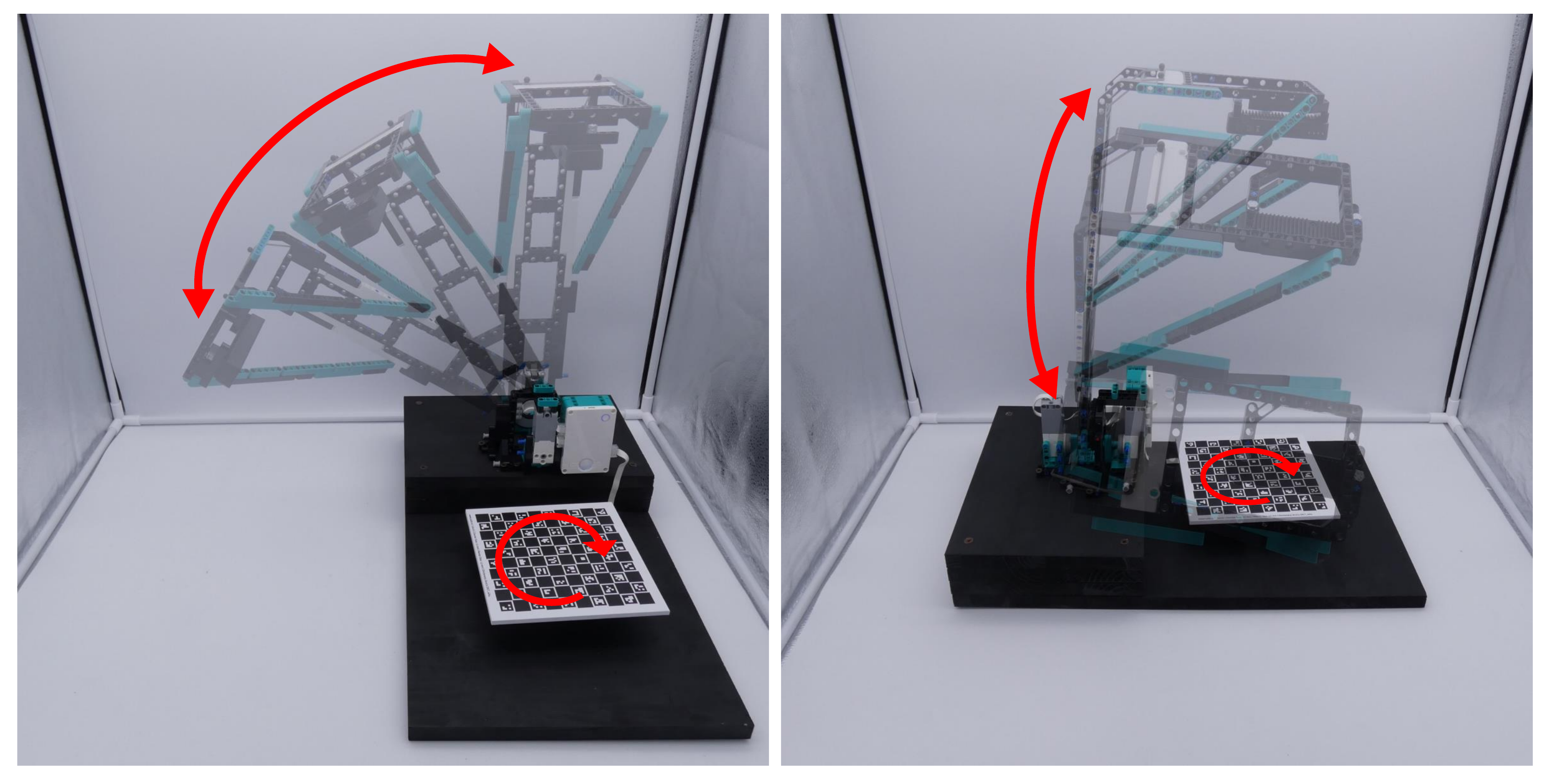}
    \caption{\textbf{The scan station.} Front and side view of our platform, with rotating angles over-imposed in red.}
    \label{fig:scan_station}
\end{figure}

The \textit{scan station} (see Fig. \ref{fig:scan_station}) that we use to generate the dataset has been built using the Lego Mindstorm toolkit (code 51515)\footnote{\url{https://www.lego.com/product/robot-inventor-51515}} and mounts an OpenCV Oak-D Lite camera\footnote{\url{https://docs.luxonis.com/projects/hardware/en/latest/pages/DM9095.html}} to collect images.
The system is composed  of a rotating base, where the object is placed during scanning, and a robotic arm holding the camera over the base. {Acquisitions are carried out inside a light box, in order to minimize effects due to shadows.}
The base and the arm are fixed on a shared structure, the latter being placed on a higher level with respect to the base, so as to allow for capturing high objects entirely. 

The arm has been built using two Lego motors (id: 6299646)\footnote{\url{https://www.lego.com/en-us/product/medium-angular-motor-88018}} connected in series to a gearbox, which holds the arm.
We use two motors and a gearbox to deploy more mechanical torque, since the arm and the camera are too heavy for the single motors alone.

The base is driven by a single, additional Lego motor, with a ChArUco board fixed on top of it, which is used to compute the camera pose for each acquired image. This is achieved by calibrating both the intrinsic and extrinsic parameters of the camera based on the ChArUco marker and the standard algorithm implemented using the functionalities made available by the OpenCV library\footnote{\url{https://docs.opencv.org/3.4/da/d13/tutorial_aruco_calibration.html}}.

To acquire images from poses that are evenly distributed on the hemisphere around the object to be scanned, the arm descends from its initial position, located vertically over the base (zenith angle $\sim$20°), to its final position, located horizontally with respect to the base (zenith angle $\sim$75°), performing sixteen total steps. After each descending step, the arm motors are stopped to hold the position, while the base performs two complete rotations (720°), to ensure a dense set of acquisitions.
During the whole process, the OAKD-Lite camera records images at 30 FPS frequency and $1440\times1080$ resolution.
The scan station has been programmed in python using the API of the Lego Mindstorm Desktop app and is controlled via bluetooth connection.

Combining the two degrees of freedom given by the arm and the rotating table enables to collect images all around the scanned object with very low effort, as well as to implement our scan station with an hardware budget resulting lower than 500\$.

\begin{figure}[t]
    \centering
    \begin{tabular}{cc}
        \includegraphics[width=0.22\textwidth]{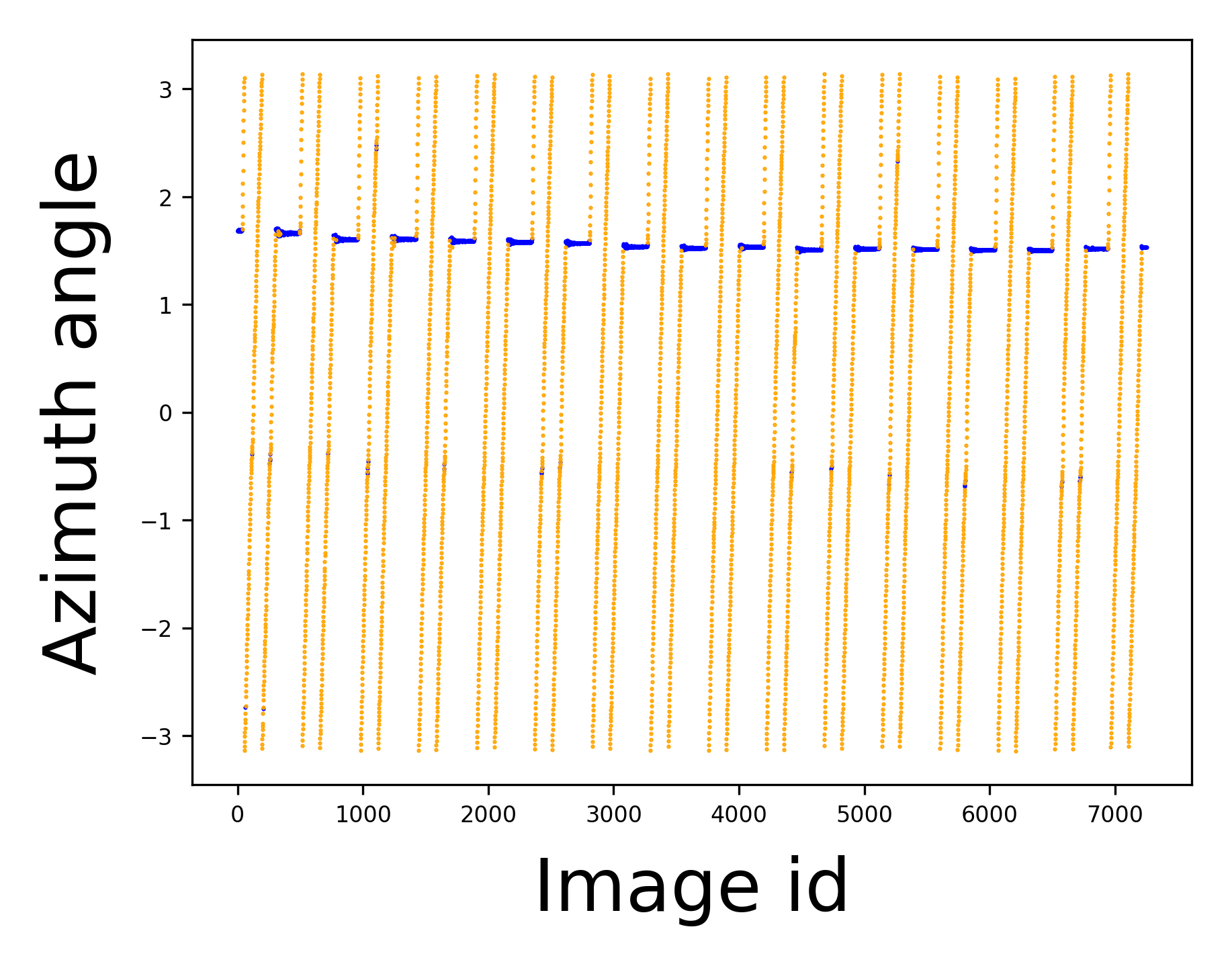}
        &
        \includegraphics[width=0.22\textwidth]{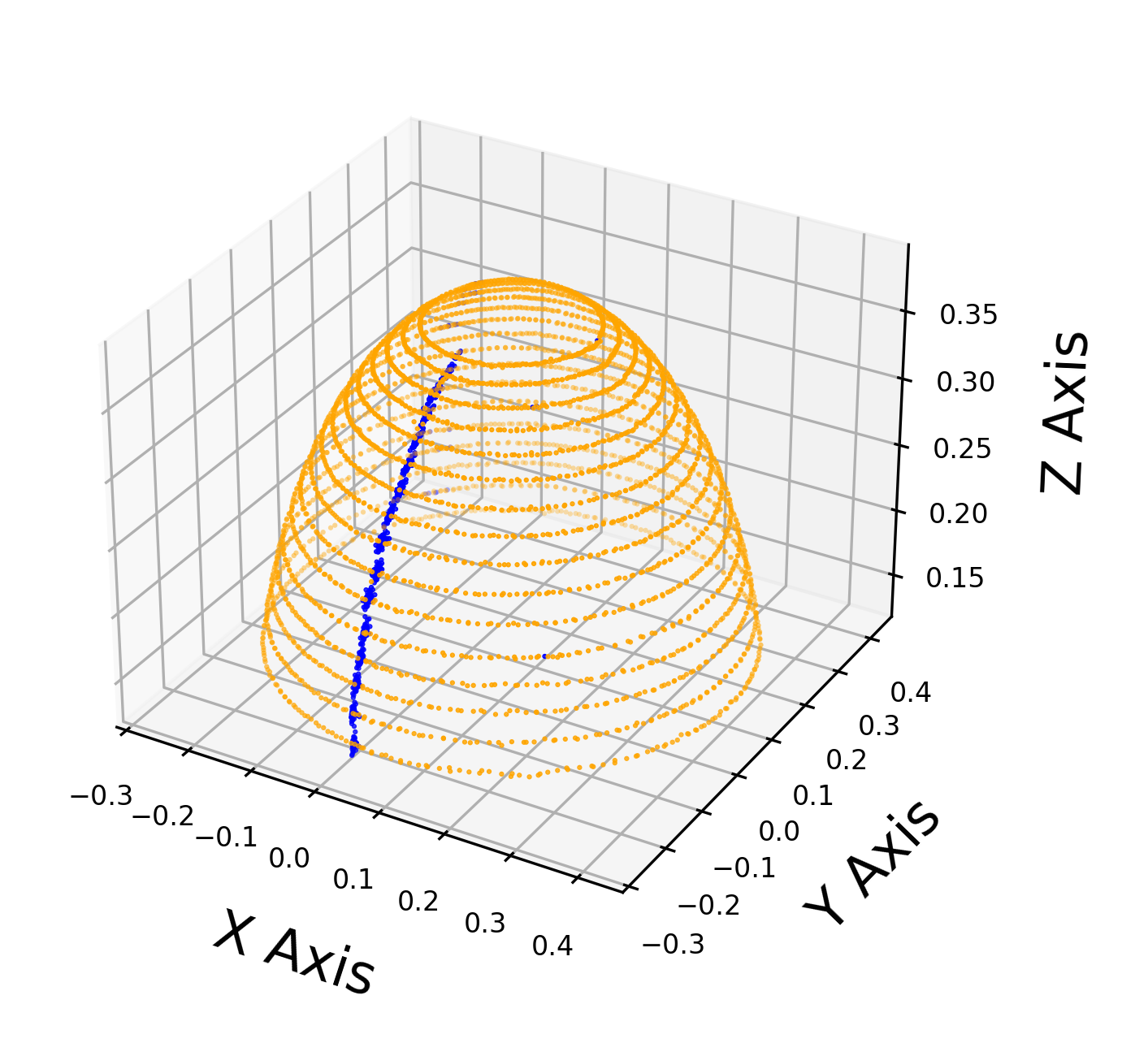} \\
    \end{tabular}

    \caption{\textbf{Filtering step.} For any collected image, we show the azimuth difference with respect to the previous one (left) and its position on the hemisphere over the object (right). We split the set of collected images into filtered (blue) and remaining ones (orange).
    }
    \label{fig:azimuth}
\end{figure}

\subsection{Dataset filtering}
\label{sec:filtering}
After a complete scanning cycle, we obtain roughly 9000 images. As images are acquired throughout the whole cycle,  some of them are captured during arm descent, i.e., the step towards the following zenith angle. This causes strong, undesired oscillation of the scan station, with consequent acquisition of several images which are blurred or out from the main trajectory.
A first cleaning step consists in removing such images, keeping only the ones obtained when the arm is not moving and the base is rotating.
We observe that the rotation of the base can be detected by computing the azimuth angle of the camera pose in each image and detecting the intervals where the angle between subsequent images is increasing. 
Thus, we discard every image whose azimuth angle differs from the previous one by less than a fixed threshold, set to 1.15°. 
Fig. \ref{fig:azimuth} shows, for an entire scanning cycle, the filtered (blue) and kept  (orange) images. We can notice how selecting the acquisitions  with smaller azimuth difference (left) effectively removes the images collected during arm descent (right).

\begin{figure}
    \centering
    \includegraphics[width=0.45\textwidth]{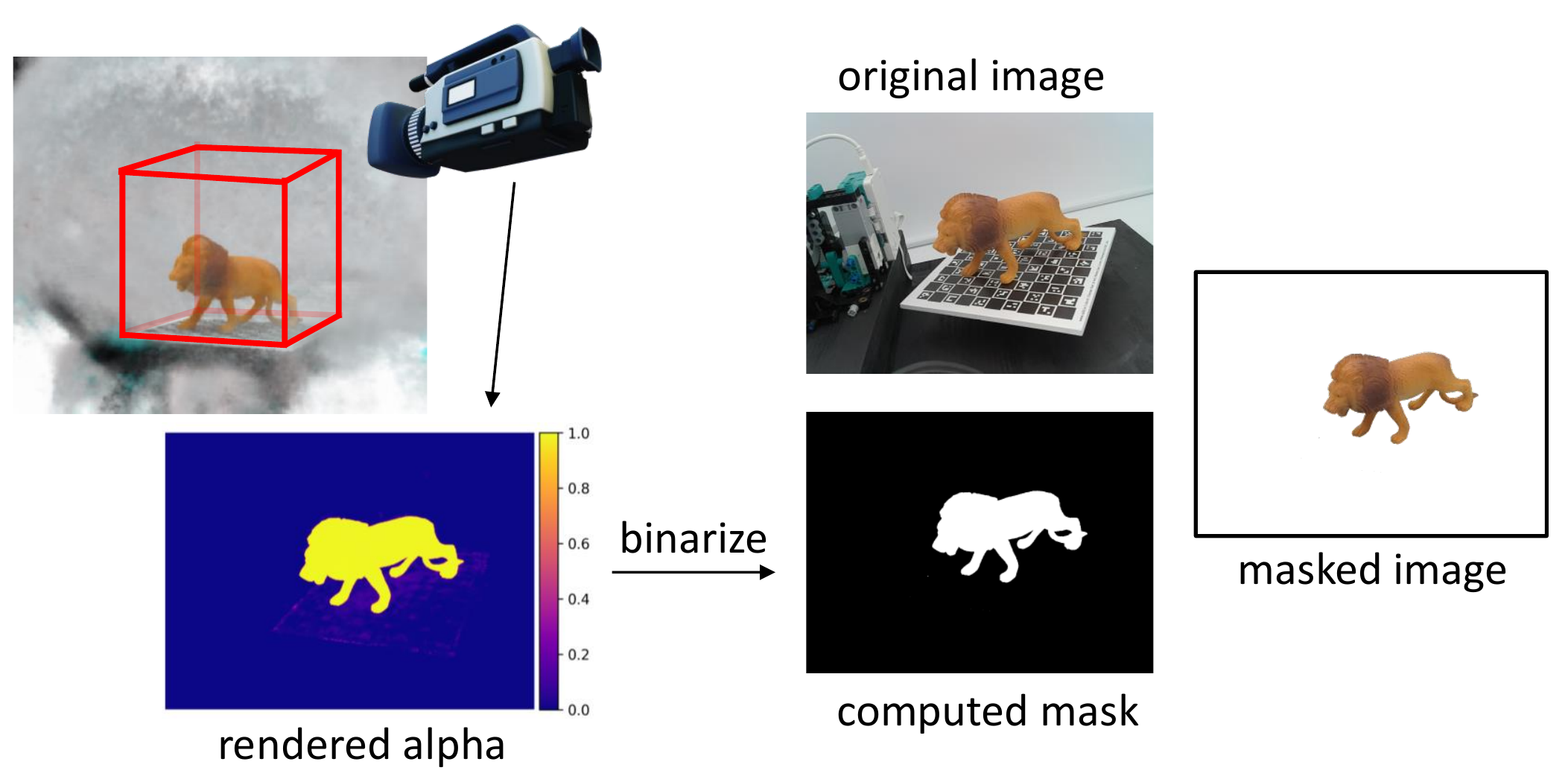}
    \caption{\textbf{Masking procedure.} We train Instant-NGP~\cite{mueller2022instant} by placing the rendering bounding-box over the ChArUco board, so as to remove the background and obtain a mask to be applied to the real image.}
    \label{fig:masking}
\end{figure}

\subsection{Background masking}
In our pipeline, we achieve the motion of the camera around the object by properly moving the scan station arm and  rotating the base on which the object is placed. This procedure presents a major side effect: the background is not coherent with the computed camera poses, since it remains still during the acquisition of the images. For this reason -- and also to obtain more pleasant images featuring only the scanned object -- we   mask out the background.

Purposely, we exploit a neural rendering framework. First, we train Instant-NGP~\cite{mueller2022instant} on the acquired images, which include  the background.
Then, we use Instant-NGP  to render new images from the same poses as the original images, defining the rendering volume to fit the ChArUco marker dimensions in order to crop out the incoherent background (Fig. \ref{fig:masking}, top left). In particular, the rendering volume is placed above the scan station base with a small offset on the Z axis so as to remove the ChArUco marker from the rendered image. This allows us to obtain rendered images featuring  the object on a black background.
Then, we binarize the rendered images based on the alpha values (i.e., density) of the pixels (Fig. \ref{fig:masking}, bottom left) to generate the desired masks (Fig. \ref{fig:masking}, bottom right). These masks are applied to the original images in order to remove both the background and the scan station base, leaving the object alone in the final images provided  by our scan station (rightmost picture in  Fig. \ref{fig:masking}). 

\begin{figure}
    \centering
    \includegraphics[width=0.48\textwidth]{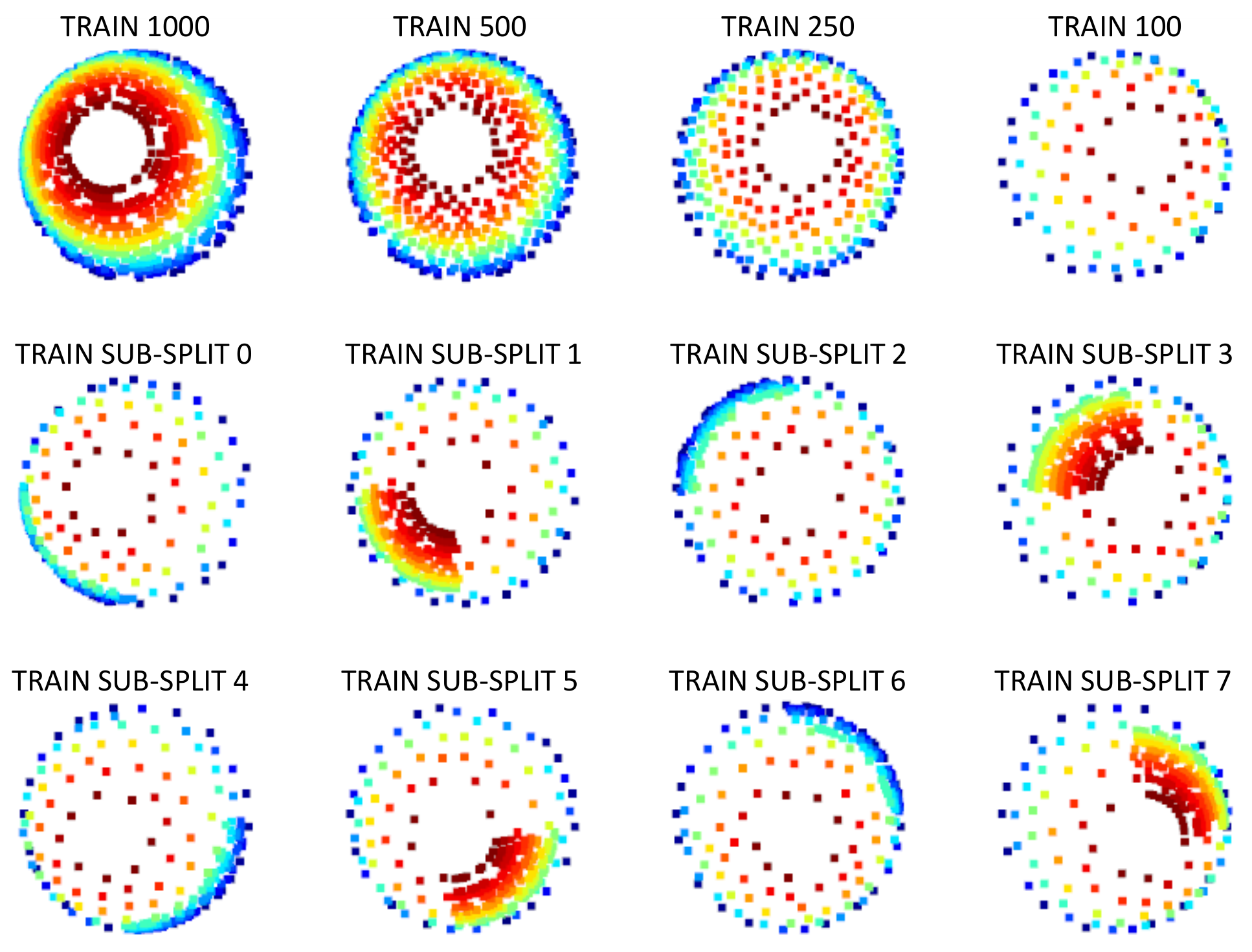}
    \caption{\textbf{Overview of the dataset splits.} On the first row, evenly sampled splits with varying density. On the second and third rows, eight sub-splits with densely localized acquisitions. Point are colored according to the Z coordinate, for a better visualization of their 3D position.}
    \label{fig:train_poses}
\end{figure}

\input{tables/overall}

\subsection{Dataset organization and splitting}
\label{sec:splitting}
Once the undesired frames have been removed and the remaining ones have been properly masked to remove the background and the scan station, we first divide each  acquired sequence into three macro-splits, namely \textit{Train}, \textit{Val} and \textit{Test}, so that they contain  1000, 500 and 500 images, respectively.
We will release the Train and the Val splits publicly, while we will keep private the Test split in order to enable a fair evaluation of the submissions that other researchers would be willing to upload on the benchmark website. 
For each split, we obtain images taken from positions evenly scattered on the hemisphere above the object by applying the Farthest Point Sampling algorithm \cite{moenning2003fast} to the 3D positions from which the images were captured.

From the 1000 images of the Train macro-split, we sample 3 smaller training splits, containing 500, 250 and 100 images,  captured uniformly from the whole hemisphere, as shown in Fig. \ref{fig:train_poses} first row. 
These additional training splits are designed to compare the performances of NeRF algorithms when trained on splits with different number of images.

Moreover, every Train/Val/Test macro-split is used to obtain eight additional sub-splits, each containing images acquired more densely in a specific region and only a small portion of images taken from positions scattered across the whole hemisphere (Fig. \ref{fig:train_poses}, second and third rows).
{Specifically, we first divide the hemisphere into eight sub-regions, by splitting each range of the X, Y and Z axes in two. Then, sub-splits are sampled from the 1000/500/500 Train/Val/Test images, by retaining all the images collected from viewpoints in the sub-regions ($\sim$ 120/60/60, with small fluctuations depending on the selected region), together with 10\% additional frames randomly sampled from the remaining portion of the hemisphere ($\sim$ 80/40/40).}

We designed these sub-splits to investigate on the performance of different NeRF proposals when the training set is characterized by  an uneven spatial distribution of vantage points and, thus, foster future research in this direction.

\subsection{Scan time and number of objects}

The pipeline sketched so far allows for effortless scanning of a large amount of objects. Specifically, an entire acquisition cycle requires about 5 minutes to collect roughly 9000 images, reduced to about 4000 after the filtering step described in Section \ref{sec:filtering}. At the time of writing, the ScanNeRF dataset counts 35 real objects over which we evaluate the performance of modern NeRF frameworks, as reported in the next Section. Moreover, we plan to scale up our dataset to hundreds (or even thousands!) of objects and distribute the associated Train/Val splits through the benchmark website. 

\section{Experiments}

In this section, we conduct experiments on our novel ScanNeRF dataset. Specifically, we run three modern and efficient NeRF frameworks \cite{sun2021direct,yu2021plenoxels,mueller2022instant} on the splits we have designed, so as to investigate on  how they perform when varying the density and amount of training images, as well as how they behave with images being densely acquired only from a specific region around the scanned object.

\input{tables/splits}

\subsection{Evaluated frameworks and settings}
We briefly introduce the methods involved in our experiments. The three of them have been selected for our evaluation because of their speed both at training and rendering time. In our opinion, such  efficiency makes these methods prominent for future advances in the field.

\textbf{DVGO \cite{sun2021direct}.} This framework mixes the implicit representation learned by means of MLPs with explicit ones -- i.e., voxel grids -- to model density and appearance. This allows for training a NeRF in roughly 15 minutes.

\textbf{Plenoxels \cite{yu2021plenoxels}.} A voxel grid is diretly optimized by this method, getting rid of any neural network. Spherical harmonics are used to model view-dependent RGB values. Training time for a single scene takes about 10 minutes.

\textbf{Instant-NGP \cite{mueller2022instant}.} This framework deploys a multi-resolution hash table of trainable feature vectors, allowing the use of a much smaller neural network and achieving faster convergence. For a single training, approximately 1 minute is enough to reach high-quality renderings.

\textbf{Training setups.} For each method, we run experiments using the official code released by the authors, keeping the same default hyper-parameters defined in the source code during training except i) for Instant-NGP, for which we reduced the amount of training step from 100K to 10K without any loss of final rendering quality, and ii) for DVGO, where we train and render half resolution images for the sake of memory constraints. In our evaluation, we trained 420 instances for each model (140 for evenly distributed acquisitions, 280 for densely localized splits). Each training is performed on a single NVIDIA 3090 RTX GPU, requiring a total of about 175 hours/GPU for training.

\textbf{Evaluation metrics.} To assess the quality of the rendered images, we compute the Peak Signal Noise Ratio (PSNR) between the rendered ($\hat{x}$) and real test ($x$) images: 

\begin{equation}
    \text{PSNR}(\hat{x},x) = -10\log_{10}(x-\hat{x})^2.
\end{equation}

\subsection{Experiments on evenly distributed acquisitions}

We start by training and evaluating the three methods when dealing with evenly distributed images taken from all around the hemisphere over the scanned object. Tab. \ref{tab:overall_results} collects experiments over 35 objects scanned by our scan station. From left to right, we report the results on the evenly distributed images of the test split achieved by training on the 1000, 500, 250 and 100 images training splits, respectively.
We can notice how all the three NeRF variants excel when trained on 1000 images, always achieving more than 30 PSNR. In general, Instant-NGP yields higher rendering quality compared to Plenoxels, while DVGO produces very good results as well, although not directly comparable with the other methods because of the limiting requirement to work with  half resolution images. When gradually reducing the density of the training images  to 500, 250 and 100, we can notice different effects on the three frameworks. Instant-NGP achieves almost unaltered quality of the rendered images, DVGO suffers a moderate drop in terms of PSNR (about 2 points when trained on the smallest training set), while Plenoxels seems to suffer the highest drop of render quality, falling to about 20 PSNR when trained with 100 images only.

According to this benchmark, Instant-NGP seems the best choice at the time of writing, thanks to its extremely fast training and rendering speed, its overall high quality and its robustness to decreasing amounts of training images.

\begin{figure*}[t]
    \centering
    \scalebox{1}{
    \begin{tabular}{cccc}
        GT & DVGO* & Plenoxels & Instant-NGP\\
        \includegraphics[width=0.1\textwidth]{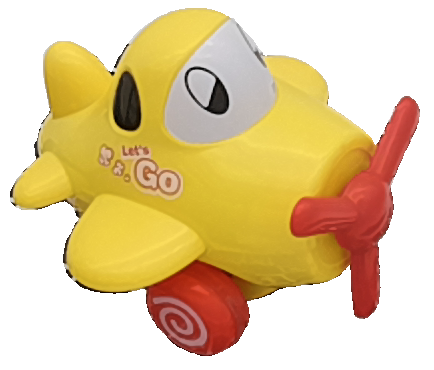}
        &
        \includegraphics[width=0.1\textwidth]{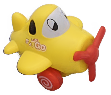}
        &
        \includegraphics[width=0.1\textwidth]{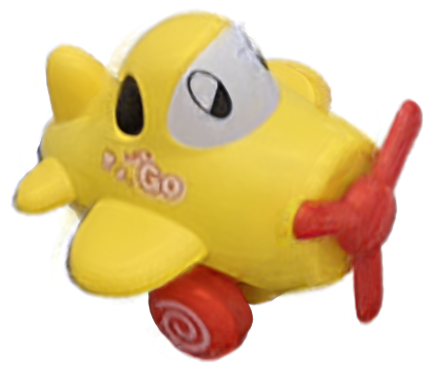}
        &
        \includegraphics[width=0.1\textwidth]{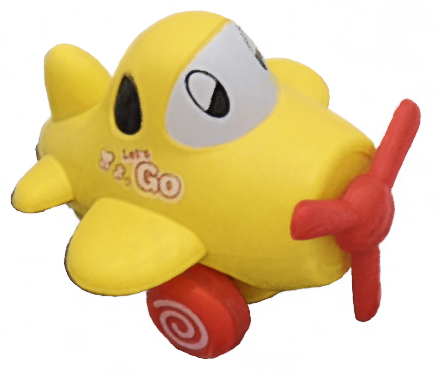}\\
        \includegraphics[width=0.18\textwidth]{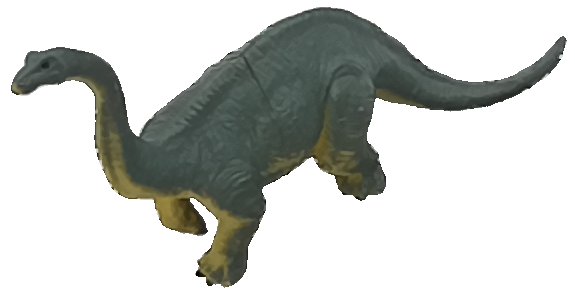}
        &
        \includegraphics[width=0.18\textwidth]{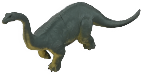}
        &
        \includegraphics[width=0.18\textwidth]{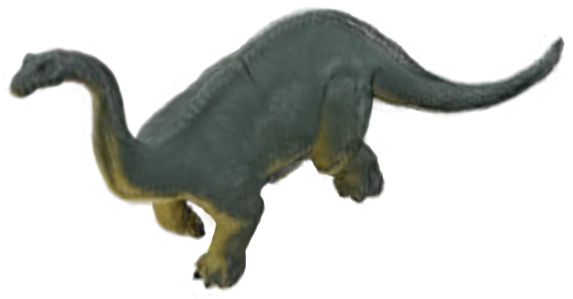}
        &
        \includegraphics[width=0.18\textwidth]{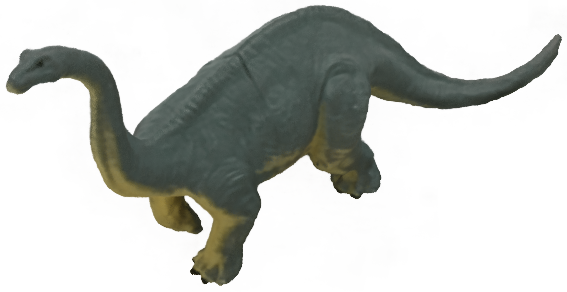}\\
        \includegraphics[width=0.12\textwidth]{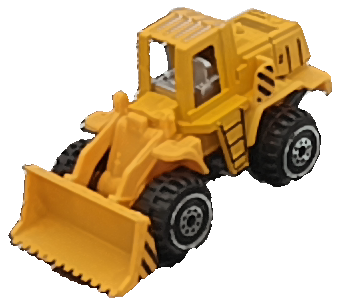}
        &
        \includegraphics[width=0.12\textwidth]{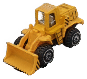}
        &
        \includegraphics[width=0.12\textwidth]{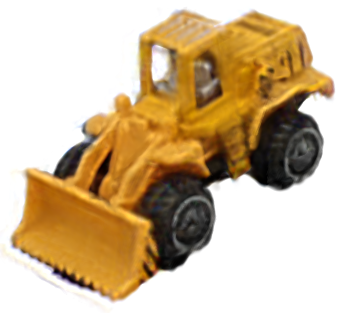}
        &
        \includegraphics[width=0.12\textwidth]{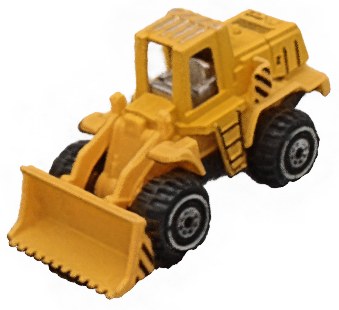}\\
        \includegraphics[width=0.1\textwidth]{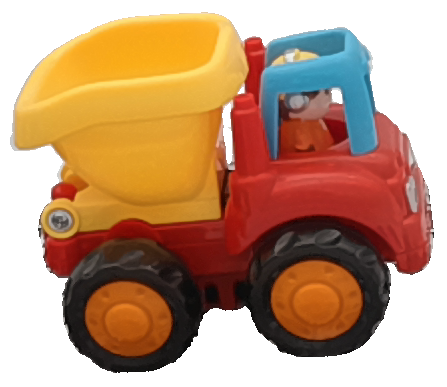}
        &
        \includegraphics[width=0.1\textwidth]{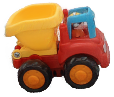}
        &
        \includegraphics[width=0.1\textwidth]{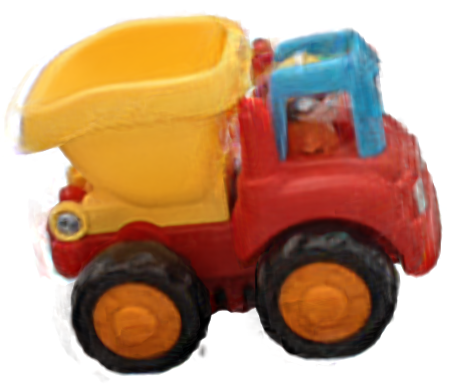}
        &
        \includegraphics[width=0.1\textwidth]{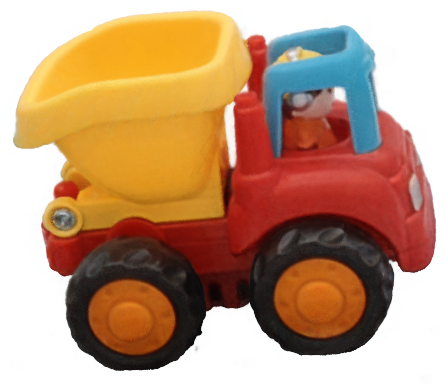}\\
        \includegraphics[width=0.1\textwidth]{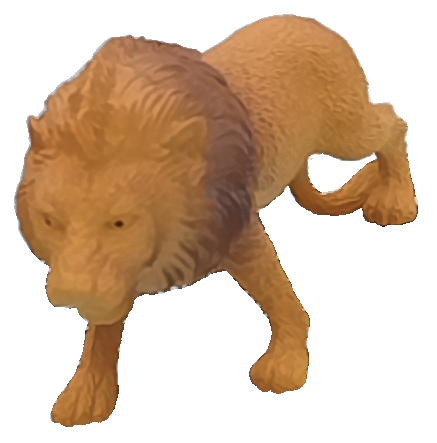}
        &
        \includegraphics[width=0.1\textwidth]{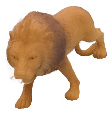}
        &
        \includegraphics[width=0.1\textwidth]{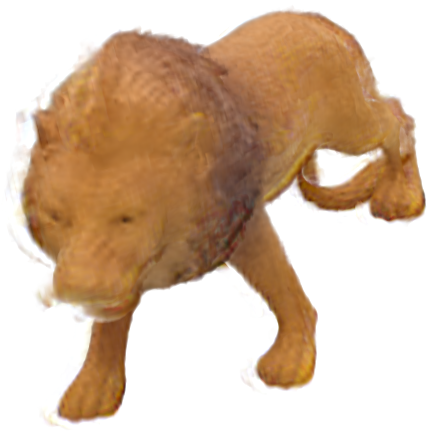}
        &
        \includegraphics[width=0.1\textwidth]{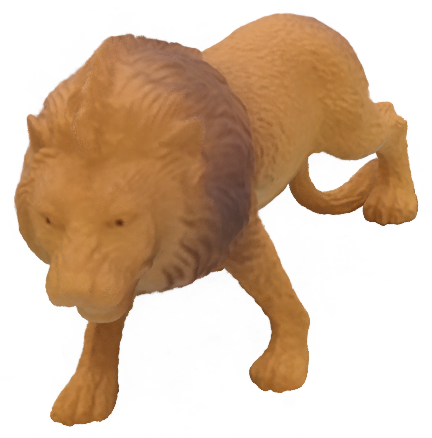}\\
        \includegraphics[width=0.1\textwidth]{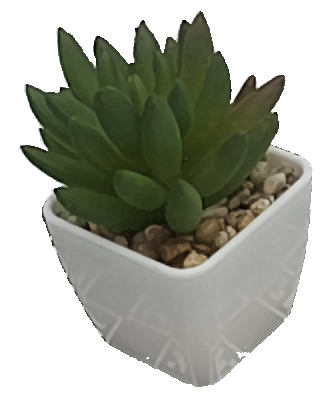}
        &
        \includegraphics[width=0.1\textwidth]{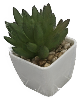}
        &
        \includegraphics[width=0.1\textwidth]{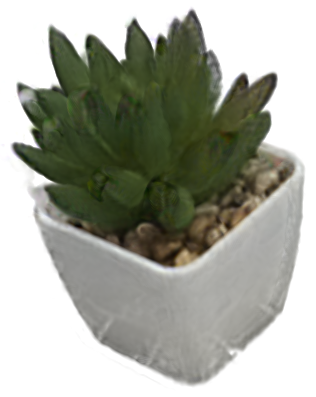}
        &
        \includegraphics[width=0.1\textwidth]{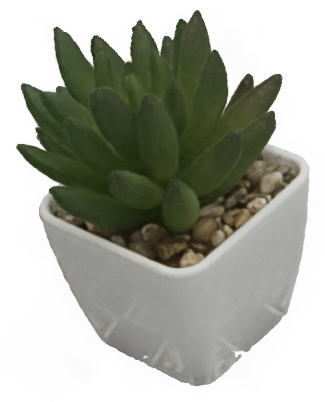}\\
        \includegraphics[width=0.1\textwidth]{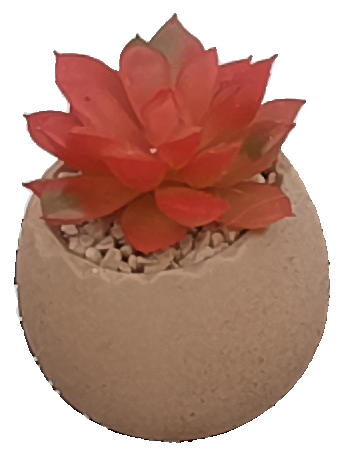}
        &
        \includegraphics[width=0.1\textwidth]{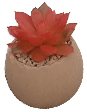}
        &
        \includegraphics[width=0.1\textwidth]{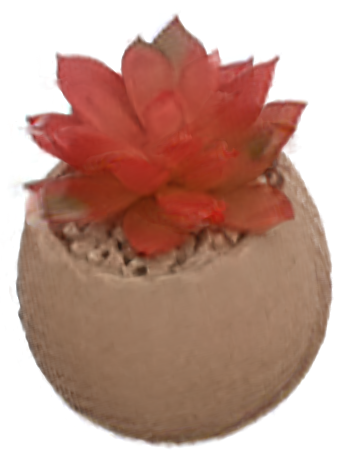}
        &
        \includegraphics[width=0.1\textwidth]{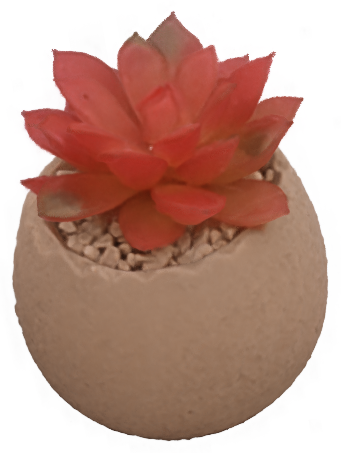}\\
        \includegraphics[width=0.11\textwidth]{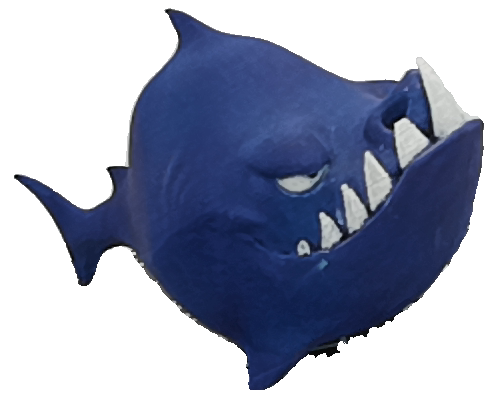}
        &
        \includegraphics[width=0.11\textwidth]{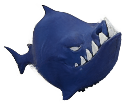}
        &
        \includegraphics[width=0.11\textwidth]{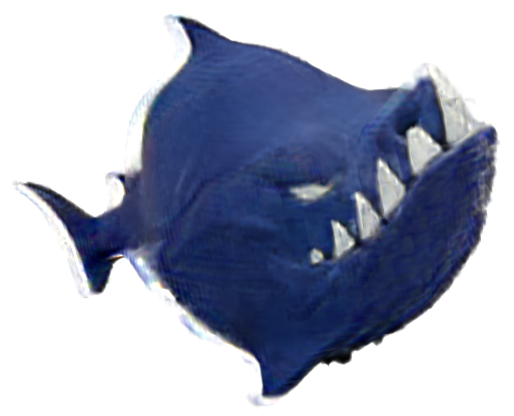}
        &
        \includegraphics[width=0.11\textwidth]{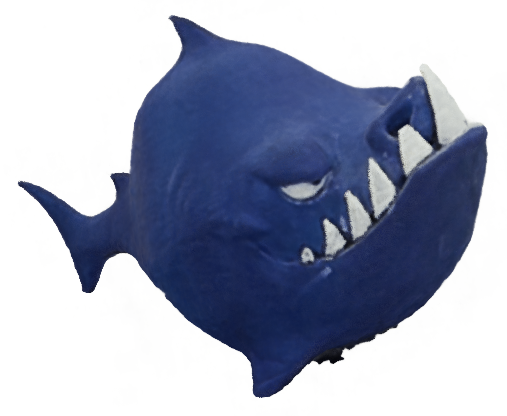}\\
    \end{tabular}}
    \caption{\textbf{Qualitative results obtained by training with 1000 images.} From left to right: ground-truth, image rendered from DVGO (half resolution), image rendered from Plenoxels, image rendered from Instant-NGP.}
    \label{fig:qualitatives}
\end{figure*}

Fig. \ref{fig:qualitatives} shows some renderings obtained by DVGO, Plenoxels and Instant-NGP when trained on 1000 images.

\subsection{Experiments on densely localized acquisitions}
After experimenting on evenly distributed acquisitions, we focus on the densely localized ones. The goal of this experiment is to stress the capability of NeRF algorithms to generate novel views from positions all over the hemisphere, after training on images captured mainly from a localized region of the space, with just few samples evenly distributed on the hemisphere.

We adopt the following protocol: for every object of our dataset, we perform eight trainings for each of the three selected NeRF algorithms (one training for every train sub-split described in Sec. \ref{sec:splitting}). Then, starting from each training, we test the three algorithms on all the eight test sub-split, performing a total of 64 evaluations for each object.

Tab. \ref{tab:splits_results} reports the results of this experiment for each selected NeRF method, averaging them on the 35 objects scanned by our framework.
It is possible to observe that, as expected, all the methodologies obtain good PSNR scores ($>$30) when trained and tested on the same sub-split (i.e., on images acquired from positions with the same distribution over the hemisphere). However, when tested on sub-splits coming from dense acquisition different from the training ones, their behavior is different from case to case. Plenoxels suffers significantly from this setting, with a PSNR drop up to 8 points that leads to poor results ($\sim$22 PSNR). DVGO, instead, appears to be more robust, with a PSNR drop inferior to 4 points. Instant-NGP, finally, seems to be the more resilient to the described stress test, with a PSNR drop of just 1 point in the worst cases.

We conjecture that the superior performances achieved by DVGO and Instant-NGP wrt Plenoxels can be explained considering that the former two methods rely on a MLP which is not present in the latter. This component can probably learn strong biases from the few evenly scattered samples, which help DVGO and Instant-NGP to generalize to (almost) unseen regions of the hemisphere.

\section{Conclusion}

In this paper, we have introduced ScanNeRF, a scalable benchmark for neural radiance fields and, in general, neural rendering frameworks. 
ScanNeRF consists of a simple, yet effective hardware/software pipeline allowing for collecting thousands images of an object effortlessly and in a few minutes. Our platform results ideal to scan a multitude of different objects, which together build up the ScanNeRF benchmark, a novel dataset made of 35 scenes counting thousands of images each.
In our experiments, we stressed the potentialities of modern NeRF frameworks \cite{sun2021direct,yu2021plenoxels,mueller2022instant} under different settings thanks to the peculiar training/validation/testing splits made available by ScanNeRF, highlighting some new challenges for the community to face.
We believe ScanNeRF will play a role in fostering the research in neural radiance fields frameworks.

{\small
\bibliographystyle{ieee_fullname}
\bibliography{egbib}
}

\end{document}

%% file: tables/overall.tex
\begin{table*}[]
    \centering
    \scalebox{0.65}{
    \begin{tabular}{l;c|c|c;c|c|c;c|c|c;c|c|c}
        \multicolumn{1}{c;}{ } & \multicolumn{3}{c;}{1000 training images} & \multicolumn{3}{c;}{500 training images} & \multicolumn{3}{c;}{250 training images} & \multicolumn{3}{c}{100 training images}\\
        \hline
        Scene & DVGO* & Plenoxels & Instant-NGP & DVGO* & Plenoxels & Instant-NGP & DVGO* & Plenoxels & Instant-NGP & DVGO* & Plenoxels & Instant-NGP\\
        \hline
        airplane1&38.90&34.59&37.14&38.97&33.49&36.40&38.41&27.44&37.57&36.69&22.81&37.30\\
        airplane2&39.82&35.21&37.86&39.85&33.69&38.38&39.46&27.21&37.61&37.60&23.36&37.44\\
        brontosaurus&41.56&34.74&39.95&41.46&30.18&39.99&40.76&24.67&39.93&38.62&20.43&39.96\\
        bulldozer1&35.84&32.05&34.99&35.95&29.78&34.72&35.70&23.68&34.90&34.05&19.34&34.72\\
        bulldozer2&39.16&34.21&38.12&38.96&34.33&37.65&37.96&32.45&38.30&36.12&26.40&38.09\\
        cheetah&37.86&33.35&35.68&37.87&32.47&35.24&37.64&29.54&21.82&36.09&23.49&35.59\\
        dumptruck1&37.93&33.90&36.61&37.93&32.41&36.78&37.44&27.14&36.60&35.63&22.01&36.65\\
        dumptruck2&41.34&35.45&39.96&41.01&34.16&39.44&40.00&30.20&38.82&38.01&25.57&39.93\\
        elephant&38.62&32.11&36.49&38.65&25.10&36.21&38.25&21.04&34.65&36.42&18.06&36.01\\
        excavator&40.87&35.23&38.65&40.65&35.33&39.59&39.82&33.74&38.48&37.83&26.90&39.77\\
        forklift&37.95&32.99&37.82&37.71&33.09&38.22&36.63&32.13&37.68&34.59&25.87&37.80\\
        giraffe&36.67&32.38&34.42&36.72&31.25&34.54&36.45&26.61&34.65&34.78&21.97&34.26\\
        helicopter1&39.77&35.52&37.71&39.73&33.35&36.84&39.29&27.55&37.57&37.56&22.81&36.98\\
        helicopter2&38.05&33.68&36.46&38.11&32.30&36.93&37.66&26.96&36.69&35.97&21.67&36.43\\
        lego&34.52&30.42&33.92&34.58&26.32&33.79&34.33&22.15&33.88&32.78&19.44&33.79\\
        lion&39.16&33.50&38.21&39.16&26.41&38.24&38.73&22.20&37.47&36.89&19.33&34.91\\
        plant1&40.31&34.41&37.21&40.34&28.29&37.23&39.72&22.72&37.42&37.44&19.99&37.03\\
        plant2&42.19&36.61&38.86&42.18&34.07&38.98&41.42&27.38&38.38&39.35&23.01&27.53\\
        plant3&33.63&29.33&33.81&33.58&24.17&34.08&33.11&20.49&34.21&30.47&18.46&33.18\\
        plant4&38.08&32.94&36.43&37.97&29.15&36.55&37.71&25.51&36.97&35.86&22.15&36.79\\
        plant5&39.10&34.30&38.11&39.06&28.02&36.64&38.48&24.01&37.18&36.28&20.79&37.99\\
        plant6&36.76&30.87&34.25&36.84&25.30&35.19&36.46&21.12&35.15&34.51&19.13&35.05\\
        plant7&37.15&31.87&35.57&37.16&26.55&35.43&36.64&20.62&35.50&34.85&18.98&35.36\\
        plant8&39.04&33.47&36.68&39.04&28.13&36.74&38.46&22.06&36.61&36.36&19.93&36.34\\
        plant9&40.05&33.79&37.52&40.07&27.44&37.39&39.36&22.03&37.44&37.42&19.57&37.51\\
        roadroller&39.96&34.66&39.18&39.62&34.59&39.66&38.84&33.46&38.94&36.61&27.28&39.37\\
        shark&39.95&32.88&38.33&39.88&25.31&38.44&39.25&19.98&38.15&37.00&17.78&38.28\\
        spinosaurus&40.86&34.96&39.31&40.88&32.73&39.09&40.44&25.81&39.32&38.71&21.74&39.21\\
        stegosaurus&39.07&33.89&38.60&39.25&29.32&37.96&38.82&25.22&38.36&37.37&22.47&38.52\\
        tiger&37.67&32.87&36.41&37.26&30.20&36.38&37.36&24.65&36.39&35.46&20.44&35.95\\
        tractor&34.02&30.55&33.51&34.10&28.67&33.88&33.87&23.34&33.31&32.42&19.32&33.73\\
        trex&37.97&32.99&37.82&38.11&29.12&37.91&37.74&22.46&37.49&35.70&18.88&38.03\\
        triceratops&41.56&35.89&39.31&41.52&32.50&40.04&40.97&25.91&39.74&39.19&22.69&39.80\\
        truck&37.70&33.67&36.36&37.68&32.80&36.64&37.30&27.53&36.66&35.67&22.44&36.50\\
        zebra&35.06&30.32&33.49&35.10&30.32&33.32&34.84&29.71&33.12&33.63&26.39&33.12\\
        \hline
        avg&38.52&33.42&36.99&38.48&30.30&36.99&37.98&25.68&36.48&36.11&21.74&36.54\\
    \end{tabular}
    }
    \caption{\textbf{Results on evenly distributed images.} We report results in terms of PSNR on the test splits of 35 scanned objects, when training with varying amount of images (from left to right, 1000, 500, 250 and 100 respectively). * indicates that DVGO has been trained and tested with half resolution images due to memory constraints.}
    \label{tab:overall_results}
\end{table*}

%% file: tables/splits.tex
\begin{table}[]
    \setlength\tabcolsep{0.5pt}
    \centering
    \scalebox{0.8}{
    \begin{tabular}{c|c|c|c|c|c|c|c|c}
        & \multicolumn{8}{c}{Test Split}\\
        \hline
        Train Split & 0 & 1 & 2 & 3 & 4 & 5 & 6 & 7\\
        \hline
        0&\gradientdvgo{39.07}&\gradientdvgo{36.54}&\gradientdvgo{36.45}&\gradientdvgo{35.81}&\gradientdvgo{36.51}&\gradientdvgo{35.76}&\gradientdvgo{36.67}&\gradientdvgo{35.97}\\
        1&\gradientdvgo{37.14}&\gradientdvgo{38.36}&\gradientdvgo{36.03}&\gradientdvgo{35.49}&\gradientdvgo{36.04}&\gradientdvgo{35.57}&\gradientdvgo{36.28}&\gradientdvgo{35.57}\\
        2&\gradientdvgo{36.74}&\gradientdvgo{36.01}&\gradientdvgo{38.91}&\gradientdvgo{36.37}&\gradientdvgo{36.22}&\gradientdvgo{35.64}&\gradientdvgo{36.86}&\gradientdvgo{36.00}\\
        3&\gradientdvgo{36.33}&\gradientdvgo{35.75}&\gradientdvgo{36.91}&\gradientdvgo{38.26}&\gradientdvgo{35.87}&\gradientdvgo{35.31}&\gradientdvgo{36.41}&\gradientdvgo{35.74}\\
        4&\gradientdvgo{36.77}&\gradientdvgo{35.95}&\gradientdvgo{36.15}&\gradientdvgo{35.65}&\gradientdvgo{38.78}&\gradientdvgo{36.34}&\gradientdvgo{36.83}&\gradientdvgo{36.07}\\
        5&\gradientdvgo{36.26}&\gradientdvgo{35.68}&\gradientdvgo{35.72}&\gradientdvgo{35.23}&\gradientdvgo{36.98}&\gradientdvgo{38.09}&\gradientdvgo{36.46}&\gradientdvgo{35.83}\\
        6&\gradientdvgo{36.58}&\gradientdvgo{35.96}&\gradientdvgo{36.42}&\gradientdvgo{35.72}&\gradientdvgo{36.57}&\gradientdvgo{35.85}&\gradientdvgo{39.20}&\gradientdvgo{36.58}\\
        7&\gradientdvgo{36.22}&\gradientdvgo{35.61}&\gradientdvgo{36.04}&\gradientdvgo{35.56}&\gradientdvgo{36.15}&\gradientdvgo{35.56}&\gradientdvgo{37.26}&\gradientdvgo{38.43}\\
        \hline
        \multicolumn{9}{c}{DVGO*}\\
    \end{tabular}}
    \\
    \scalebox{0.8}{
    \begin{tabular}{c|c|c|c|c|c|c|c|c}
        & \multicolumn{8}{c}{Test Split}\\
        \hline
        Train Split & 0 & 1 & 2 & 3 & 4 & 5 & 6 & 7\\
        \hline
        0&\gradientplx{31.05}&\gradientplx{24.74}&\gradientplx{24.68}&\gradientplx{22.37}&\gradientplx{24.91}&\gradientplx{22.55}&\gradientplx{24.46}&\gradientplx{22.27}\\
        1&\gradientplx{27.97}&\gradientplx{30.10}&\gradientplx{24.62}&\gradientplx{23.15}&\gradientplx{24.85}&\gradientplx{23.60}&\gradientplx{24.33}&\gradientplx{22.45}\\
        2&\gradientplx{25.10}&\gradientplx{22.62}&\gradientplx{31.37}&\gradientplx{25.01}&\gradientplx{24.02}&\gradientplx{21.86}&\gradientplx{25.47}&\gradientplx{22.67}\\
        3&\gradientplx{24.81}&\gradientplx{23.32}&\gradientplx{28.09}&\gradientplx{30.17}&\gradientplx{23.50}&\gradientplx{21.72}&\gradientplx{24.85}&\gradientplx{23.30}\\
        4&\gradientplx{25.16}&\gradientplx{22.56}&\gradientplx{24.09}&\gradientplx{22.00}&\gradientplx{31.17}&\gradientplx{25.22}&\gradientplx{25.47}&\gradientplx{22.84}\\
        5&\gradientplx{24.83}&\gradientplx{23.17}&\gradientplx{23.66}&\gradientplx{22.06}&\gradientplx{28.18}&\gradientplx{30.30}&\gradientplx{25.16}&\gradientplx{23.87}\\
        6&\gradientplx{24.20}&\gradientplx{22.15}&\gradientplx{24.96}&\gradientplx{22.64}&\gradientplx{24.79}&\gradientplx{22.35}&\gradientplx{31.36}&\gradientplx{25.06}\\
        7&\gradientplx{23.90}&\gradientplx{22.31}&\gradientplx{24.73}&\gradientplx{23.45}&\gradientplx{24.72}&\gradientplx{23.21}&\gradientplx{28.02}&\gradientplx{30.10}\\
        \hline
        \multicolumn{9}{c}{Plenoxels}\\
    \end{tabular}}
    \\
    \scalebox{0.8}{
    \begin{tabular}{c|c|c|c|c|c|c|c|c}
        & \multicolumn{8}{c}{Test Split}\\
        \hline
        Train Split & 0 & 1 & 2 & 3 & 4 & 5 & 6 & 7\\
        \hline
        0&\gradientinst{36.98}&\gradientinst{36.10}&\gradientinst{36.43}&\gradientinst{36.04}&\gradientinst{36.34}&\gradientinst{35.75}&\gradientinst{36.31}&\gradientinst{35.92}\\
        1&\gradientinst{36.23}&\gradientinst{36.99}&\gradientinst{36.19}&\gradientinst{36.24}&\gradientinst{36.14}&\gradientinst{35.93}&\gradientinst{36.07}&\gradientinst{35.95}\\
        2&\gradientinst{36.54}&\gradientinst{36.31}&\gradientinst{37.40}&\gradientinst{36.64}&\gradientinst{36.54}&\gradientinst{36.17}&\gradientinst{36.61}&\gradientinst{36.29}\\
        3&\gradientinst{36.17}&\gradientinst{36.18}&\gradientinst{36.53}&\gradientinst{37.26}&\gradientinst{36.19}&\gradientinst{35.94}&\gradientinst{36.23}&\gradientinst{36.21}\\
        4&\gradientinst{36.39}&\gradientinst{36.00}&\gradientinst{36.48}&\gradientinst{36.12}&\gradientinst{37.12}&\gradientinst{36.10}&\gradientinst{36.46}&\gradientinst{36.09}\\
        5&\gradientinst{36.07}&\gradientinst{36.14}&\gradientinst{36.20}&\gradientinst{36.16}&\gradientinst{36.45}&\gradientinst{36.94}&\gradientinst{36.21}&\gradientinst{36.21}\\
        6&\gradientinst{36.43}&\gradientinst{36.25}&\gradientinst{36.63}&\gradientinst{36.42}&\gradientinst{36.58}&\gradientinst{36.15}&\gradientinst{37.28}&\gradientinst{36.48}\\
        7&\gradientinst{36.17}&\gradientinst{36.11}&\gradientinst{36.39}&\gradientinst{36.36}&\gradientinst{36.31}&\gradientinst{36.15}&\gradientinst{36.50}&\gradientinst{37.20}\\
        \hline
        \multicolumn{9}{c}{Instant-NGP}\\
    \end{tabular}}
    \caption{\textbf{Results on densely localized sub-splits.} From top to bottom: DVGO (half-resolution), Plenoxels and Instant-NGP. We show results in terms of PSNR, averaged over the 35 scanned objects, for models trained on one of the eight densely localized sub-splits (rows) and tested on any of the eight sub-splits (columns). }
    \label{tab:splits_results}
\end{table}